# A generalized financial time series forecasting model based on automatic feature engineering using genetic algorithms and support vector machine

Norberto Ritzmann Júnior

Programa de Pós-Graduação em Informática

Pontifícia Universidade Católica do Paraná
Curitiba, Brazil
norberto.ritzmann@gmail.com

Julio Cesar Nievola

Programa de Pós-Graduação em Informática

Pontifícia Universidade Católica do Paraná
Curitiba, Brazil
nievola@ppgia.pucpr.br

*Abstract* — We propose the genetic algorithm for time window optimization, which is an embedded genetic algorithm (GA), to optimize the time window (TW) of the attributes using feature selection and support vector machine. This GA is evolved using the results of a trading simulation, and it determines the best TW for each technical indicator. An appropriate evaluation was conducted using a walk-forward trading simulation, and the trained model was verified to be generalizable for forecasting other stock data. The results show that using the GA to determine the TW can improve the rate of return, leading to better prediction models than those resulting from using the default TW.

## I. INTRODUCTION

Stock market movement prediction is a dynamic and nonlinear problem, and its forecasting presents a challenge. The authors of previous studies who aimed at building a model to forecast stock market movement rooted their investigations in technical analysis features based on statistical and chart patterns of historical data. These features were normally based on a time window of a certain period, normally days, in the past. Although the value of the time window variable could completely change the model results, the authors of these studies simply fixed the time window (TW) by taking the default value of this variable for each feature. Although many authors proposed new preprocessing methods to improve their models, such as the feature selection (FS) methods [1]–[3] or discretization methods [4]–[6], none of them conducted research to determine the global TW optimum values.

This study aimed to build a trading simulation based on a generalized model to classify the next day movement. In studies based only on next day classification, a supervised classifier training using one-day ahead movement as a binary class was built, and out-of-sample instances were tested to obtain the accuracy of the results or other measurements. Kim [7] proposed the use of a support vector machine (SVM) rather than case-based reasoning or an artificial neural network (ANN) as a classifier to predict the next day movement.

Following Kim's study, many studies compared machine learning classification algorithms. For example, Ou and Wang [8] experimentally investigated 10 classification algorithms to predict the stock market movement. Their experimental results showed that the SVM and least squares SVM algorithms generated better predictive performance than the others. Kara [2] and Patel [6] determined the best parameter combination for machine learning algorithms and tested several combinations of technical indicators and preprocessing techniques for these features.

In recent studies, authors tended to combine an SVM or ANN with preprocessing techniques and sometimes used a meta-heuristics algorithm to find the optimal machine learning parameters, the ANN architecture, and a set of input features [9], [10]. Other researchers built a classifier model and used the classification output as decision support for trading stocks, taking the rate of return (RR) as the measurement. Thus, some works, such as those of Li [11], Chang [12], and Ng [3], were based on the classification of turning points. Chang [12] proposed an ANN backpropagation model based on an improved piecewise linear representation to detect real points that indicate trading opportunities. Li [11] built a GA-based threshold optimization model as a trading design. Using the same turning point detection method, Ng [3] proposed a two-step GA for FS and an ANN architecture to build a model for trading simulation. The authors built a trading algorithm to determine the turning points using a GA to minimize the localized generalization error measurement and to balance the classifications into sell, buy, and hold such that they have the same distribution. Conversely, Żbikowski [13] based his trading simulation on a modification of the SVM called the volume-weighted SVM (VW-SVM), used the real transaction volume of the stock as a penalty function and F-score FS, and tested the algorithm with a stock market simulation.

The proposed method uses a combination of a TW optimization for each technical analysis feature and an embedded FS in a GA evolved by the RR of the simulation of transactions. This paper is organized as follows. Section 2 explains the prediction models related to this work. Section 3 provides the details of the proposed model. Section 4 presents the experimental results, and Section 5 concludes the paper.

## II. SVM

SVM has been widely used with success in many studies in the stock prediction context. This classifier improves its generalization ability by constructing a model maximizing the margin $\|w\|$ dividing hyperplanes in an *n*-dimensional feature space. This maximized margin is used as the final decision boundary. An SVM binary classifier was implemented by Cortes and Vapnik [14] using the structural risk minimization principle based on statistical learning theory for controlling generalization to determine the ideal tradeoff between structural complexity and empirical risk [2].

The SVM works by mapping the input vectors $x_i \in R^d$ (*i=1,2, ..., N*) by class label $y_i \in \{+1, -1\}$ into a high-dimensional feature space $\phi(x_i) \in H$ to build the optimal separating hyperplane. The classifier is defined by (1), and the quadratic programing problem to determine the coefficients $\alpha_i$ is defined by (2), (3), and (4) [15].

$$f(x) = sgn\left(\sum_{i=1}^{N} y_i \alpha_i \cdot k(x_j, x_i) + b\right) \quad (1)$$

Quadratic Problem:

$$Max \sum_{i=1}^{N} \alpha_i - \frac{1}{2}\sum_{i=1}^{N}\sum_{j=1}^{N} \alpha_i \alpha_j \cdot y_i y_j \cdot k(x_i, x_j) \quad (2)$$

$$Subject\ to\ 0 \leq \alpha_i \leq c \quad (3)$$

$$\sum_{i=1}^{N} \alpha_i y_i = 0, i = 1,2, \ldots, N \quad (4)$$

The tradeoff between the margin and the misclassification error is controlled by the regularization parameter *c*. $k(x_j, x_i)$ is the kernel function that performs the mapping of the input vectors into the high-dimensional space $\phi(x_i)$. The simplest kernel function is linear, but the polynomial and the Gaussian radial basis kernel function (RBF) are the most frequently used kernels. These kernel functions are shown in (5) and (6), respectively, where *d* is the degree of the polynomial kernel and $\gamma$ is the bandwidth of the Gaussian radial basis function kernel.

Polynomial Function: $k(x_i, x_j) = (x_i \cdot x_j + 1)^d \quad (5)$

Radial Basis Function: $k(x_i, x_j) = exp(-\gamma \|x_i - x_j\|)^d \quad (6)$

## III. GENETIC ALGORITHM FOR TIME WINDOW OPTIMIZATION (GATWO)

As previously stated, the main goal of GATWO is to optimize the TW variable of each technical analysis feature at the same time the FS is conducted in a GA optimization. Eventually, GATWO will have built an SVM model along with the selected features, its TW sizes, and a data scaler for each feature. This data scaler is used to scale new entries according to the normalization performed in the training phase. In every generation of the GA, an SVM prediction model is trained and evaluated with the resultant set of the features. This optimization finds the best value and the best set for the TW variables, represented by *n* in the equations presented in Table II, at the same optimization.

As conducted in this work, prediction models can use meta-heuristic algorithms, such as GAs, to maximize the accuracy or the profit of a trading simulation. This type of optimization can be programmed to determine better parameters or to optimize a certain transformation of the input features in the preprocessing stage. The GA is inspired by natural selection that evolves a population of individuals, with each one being composed of a set of genes (chromosome) as the parameters of a maximization or minimization function called the fitness function. Each iteration of the evolving process is composed of the operations of the crossover, mutation, and selection of the individuals. The selection of the individuals probabilistically retains the superior results and discards the others. In this work, the chromosome comprises two genes of each technical analysis feature: the first gene indicates the TW variable value and the second gene represents the selection variable. The format of the chromosome is presented in Table III. The formed chromosome is submitted to the fitness function, which is the maximization of the rate of return of the trading algorithm that evaluates the performance of the model over three stocks. The fitness evaluation function is presented in (9), and the trading algorithm is explained in detail in Section 3.A. This algorithm uses the SVM trained model in the evaluation to determine whether to buy, hold, and sell. For each sell event, the total profit (*TP*) is calculated as presented in (7):

$$TP(x) = \sum_{t=0}^{NS}[A_{sell(t)} * P_{sell(t)} - A_{buy} * P_{buy}] - NT * tcost \quad (7)$$

where *A* and *P* are the amount of stocks and the price per stock, and *NT*, *NS*, and *tcost* are the total number of transactions, number of sales, and the cost per transaction, respectively. The cost per transaction is fixed at $5.00 because this is the transaction price charged by low-cost brokers.

Moreover, the rate of return *RR* is calculated, as expressed in (8), where *Inv* is the start amount invested [11].

$$RR(x) = \frac{TP(x)}{Inv} * 100 \quad (8)$$

The *RR* average of the datasets is returned as the fitness function for the GA process. The GA takes this *RR* average as the evolving score for the maximization of the profits, as shown in (9), where $RR_s$ is the rate of return of the dataset from index *s*, and *n* is the number of training datasets present in the building model phase. The datasets are explained at Section 3.A.

$$GATWO(x) = \max\left(\frac{1}{n}\sum_{s=0}^{n} RR_s\right) \quad (9)$$

*A. Data Preparation*

In this study, the stock dataset for training was different from that for evaluation to attain the purpose of the highly generalized trained model, which can predict from any other stock dataset. The chosen training daily data were Microsoft (MSFT), Nike (NKE), and Goldman Sacks (GS) from January 1, 2000 to December 31, 2004 (Yahoo Finance). Two years of stock datasets with different trends were used for the trading evaluation (Table I).

TABLE I. GENES OF GATWO

| Stock | Period | Trend |
|---|---|---|
| YHOO | 01/01/2013–12/31/2014 | Up-trend |
| FORD | 01/01/2008–12/31/2009 | Side-trend |
| JPM | 11/01/2006–10/31/2008 | Down-trend |

TABLE II. FEATURES USED IN THE STUDY

| Name of Feature | Formula |
|---|---|
| Stochastic %K (STK) | $C_t + LL_{t-n}/HH_{t-n} - LL_{t-n} \times 100$ |
| Stochastic %D (STD) | $\frac{1}{n}\sum_{i=0}^{n-1} \%K_{t-i}$ |
| Relative strength index (RSI) | $100 - \frac{100}{1 + (\sum_{i=0}^{n-1} Up_{t-i})/(\sum_{i=0}^{n-1} Dw_{t-i}/n)}$ |
| Commodity channel index (CCI) | $\frac{M_t - SM_t}{0.015 D_t}$ |
| Weighted moving average bias (WMA bias) | $C_t - \left(\sum_{i=1}^{n}(n-i)\mu_{t-i}\right)$ |
| Psychological line (PSY) | $\frac{\sum_{i=0}^{n} NUp_i}{n} \times 100$ |
| Plus directional indicator (+DI) | $\frac{\sum_{i=0}^{n} +DM_{t-i}}{\sum_{i=0}^{n} TR_{t-i}} \times 100$ |
| Minus directional indicator (-DI) | $\frac{\sum_{i=0}^{n} -DM_{t-i}}{\sum_{i=0}^{n} TR_{t-i}} \times 100$ |
| Average directional movement index (ADX) | $SMA\left(\frac{|+DI - (-DI)|}{+DI + (-DI)}\right) \times 100$ |
| Aroon Up | $[(n - n_{LastHH})/n] \times 100$ |

Table II presents the 10 technical indicators extracted from the data, where $C_t$, $L_t$, $H_t$, and $V_t$ are the closing price, low price, high price, and volume at time $t$, respectively. $LL_{t-n}$ and $HH_{t-n}$ are the lowest low and the highest high in the preceding $n$ days before $t$. The variables $M_t$, $SM_t$, and $D_t$ are calculated by (10), (11), and (12).

$$M_t = \frac{(H_t + L_t + C_t)}{3} \quad (10)$$

$$SM_t = \frac{(\sum_{i=1}^{n} M_{t-i})}{n} \quad (11)$$

$$D_t = \frac{\sum_{i=1}^{n}|M_{t-i+1} - SM_t|}{n} \quad (12)$$

Moreover, -DM, +DM, and TR are calculated by (13), (14), and (15).

$$-DM = \begin{cases} L_{t-1} - L_t & \forall\ L_{t-1} - L_t > 0 \\ "0" & \text{otherwise} \end{cases} \quad (13)$$

$$+DM = \begin{cases} H_t - H_{t-1} & \forall\ H_t - H_{t-1} > 0 \\ "0" & \text{otherwise} \end{cases} \quad (14)$$

$$TR = \max(\{H_t - L_t, |H_t - C_{t-1}|, |L_t - C_{t-1}|\}) \quad (15)$$

$n_{LastHH}$ is the number of periods since the highest high achieved in the $n$ periods. $Up$ is the upward price change, $Dw$ is the downward price change at time $t$, and $NUp$ is the number of rising periods.

The direction of the daily change in the stock price is used as a class, which is classified as 1 for equal or higher and 0 for a lower next day price as defined in (16).

$$\text{class} = \begin{cases} "1" & \forall\ C_{t+1} \geq C_t \\ "0" & \forall\ C_{t+1} < C_t \end{cases} \quad (16)$$

*B. GA Configuration and Encoding*

This study was based on an elitist GA selection, which selects a highly scored population to keep for the next generation. About 70% of the best individuals with the highest trading simulation results were maintained for the next generation. The population size was fixed at 30 chromosomes. To replace the removed population, new individuals were generated by cloning randomly individuals. Before the subsequent round, the crossover operation was applied in 40% of the entire population, and mutation was conducted in 10% of the genes of all individuals. The range of each gene's values is provided in detail in Table III, where the TW and FS value ∈ ℕ. The stop condition was set to run until 100 rounds after the latest highest value.

TABLE III. GENES OF GATWO

| Features | TW | FS | Default TW |
|---|---|---|---|
| STK | 8–14 | 0–1 | **9** [3] |
| STD | 3–6 | 0–1 | **3** [3] |
| RSI | 5–14 | 0–1 | **6** [3] |
| PSY | 10–15 | 0–1 | 12 [16], [17] |
| WMA | 6–15 | 0–1 | 10 [18] |
| CCI | 6–15 | 0–1 | 14 [17] |
| ADX / -DI / +DI | 6–15 | 0–1 | 10 |
| Aroon Up | 19–28 | 0–1 | 25 |

FS is performed simultaneously with the TW optimization by GATWO. As shown in Table III, each technical indicator has two genes: one to control the TW and the second to control the FS. The FS value 1 means that the feature is considered in the training model; otherwise, it is not considered. This dual optimization provides to the classification a real interaction between the features during the evolution of the population. To accomplish this simultaneous optimization, the chromosome was encoded with TW and FS values.

The entire process, shown in Figure 1, is related to the resultant chromosome generated by the continuous evolving process. For each chromosome, each technical indicator is calculated using the given TW value, and then the features are normalized (details provided in Section 3.C). After this transformation, the data are used to train the SVM classifier to classify the next day movement, and the resultant classifier is used in the trading simulator. The trading simulation obtains the signal from the trained SVM classifier. Buying and selling with the three evaluation trading datasets is shown in Figure 3. The resultant average RR from these three stock simulations is used as the fitness function evaluation for the GA process and the best model is stored after each round of evolution for comparison with other studies.

*C. Data Normalization*

Classifiers based on the distance between instances, such as SVM, require a normalization of the features, which rescales each feature value. In this study, each feature was normalized using z-score normalization, given by (17), where $x'$ is the transformed value, $x$ is the original value of the feature, $\mu$ is the mean, and $\sigma$ is the standard deviation of the population for the given feature.

$$x' = \frac{(x-\mu)}{\sigma} \quad (17)$$

To simulate the transactions, new instances to be classified by the model need to be input. To use the same model with another stock dataset, the GATWO needs to continue the TW, aside from the generated SVM model, of each gene and the resultant normalization scales. As shown in (17), to transform new instances, the mean and standard deviation must be stored to scale new input data. Figure 2 shows the manner through which this process submits new data to be used.

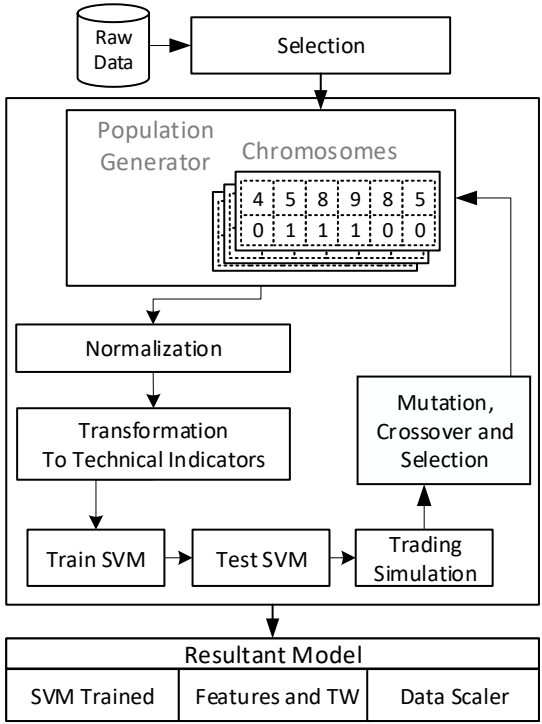

Fig. 1. GATWO model training framework.

The new raw data are input in the GATWO model, and these data are transformed into features according to each TW and selected features of the model. Then, each feature is scaled by the scaler composed of the mean and the standard deviation. Finally, these scaled data are sent to the SVM trained model. In obtaining the class, for example, the trading simulation can choose whether to buy, hold, or sell the stock, as presented in the following section.

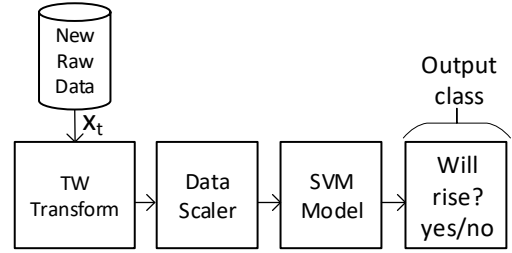

Fig. 2. GATWO prediction framework.

*D. Trading Simulation*

The algorithm uses the model built in the training step of the classifier to classify each transaction as buy, sell, or hold. The rules of the algorithm are shown in detail in Figure 3, where $f(x')$ is the resultant classification for the $x'$ instance. Only after the stock has been sold is it possible to evaluate whether the decision is profitable or not. The algorithm only buys everything possible with the available money and sells every stock; there is no mid-ground. The trading algorithm was run on the test datasets.

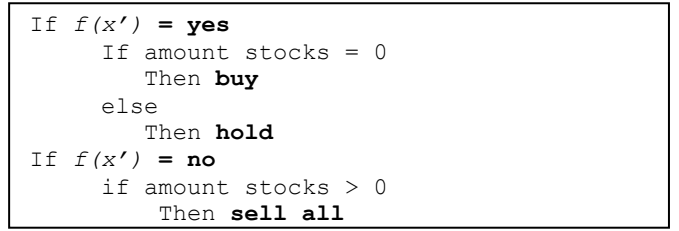

Fig. 3. Trading algorithm of the experimental results.

The experiments were conducted after running GATWO to generate the model, as presented in Figure 1, and trade was conducted according to the simulation using the new stock datasets produced by the GA process. The following were evaluated:
- *RR*, shown in (8), is used as the fitness function for the GA process.
- Maximum drawdown (*MDD*), which is used as a risk measurement, is the lowest value obtained in the transactions result. It is the worst percentage decrease between the highest and the lowest value the investor can obtain. It is calculated by using the percentage difference between the highest and the lowest account value shown in (18).

$$MDD = min\left(\frac{accumulated\ profit(x)}{\max(accumulated\ profit(x))-1}\right) \quad (18)$$

In the following section, the submission of the raw dataset to GATWO to build the model using the test stock datasets is described. In the first step, the SVM RBF configuration is tested, and the training stock data are selected.

IV. EXPERIMENTS AND DISCUSSION

The experiment begins to find the best parameters and the best training dataset, consequently building the best generalized model. After building the model, it is tested under stress and up-trend rally conditions, and the final experiments are compared with the literature results.

*A. GATWO Model Configuration*

The SVM RBF configuration tested the parameters shown in (6), namely, the parameters cost, $c$, and $\gamma$. To select the best SVM parameter configuration, GATWO was run on a grid search algorithm to determine the best composition of these two parameters. The chosen stock datasets for training were Microsoft (MSFT), Nike (NKE), Goldman Sacks (GS), and Intel (INTC) between January 2000 and December 2004. A simple grid search was chosen to find the best combination of the SVM parameters to test the values of c between 0.125 and 100 and the values of $\gamma$ between 0.1 and 10. The best combination of parameters was $c = 1$ and $\gamma = 0.25$.

As stated previously, these experiments aimed to determine the most generalized model. To accomplish this, finding the most generable training dataset among the four training datasets was necessary to determine the SVM RBF parameters $c$ and $\gamma$. To test these models, the datasets in Table I were used as test datasets. The results are presented in Table IV.

TABLE IV. RESULTS OF THE RATE OF RETURN FOR EACH STOCK DATA

| Test Data | MSFT | NKE | GS | INTC |
|---|---|---|---|---|
| JPM | 286.9 | 67.3 | 98.3 | 124.5 |
| FORD | 2437.1 | 894.5 | 1422.8 | 2537 |
| YHOO | 312.8 | 334 | 189.1 | 332.2 |
| Average | 1012.3 | 431.93 | 570.06 | 997.9 |

As shown in Table IV, the MSFT stock data were the most generalizable for training the model.

As the efficacy of the generalization of the model was considered an issue, normalization also needed to be tested because applying the correct procedure for scaling the features would result in higher yields. Until this point, z-score (17) was used as the normalization method and compared with the min–max normalization given in (19) in Table V.

$$X'_i = \frac{X_i - X_{min}}{X_{max} - X_{min}} \quad (19)$$

TABLE V. NORMALIZATION COMPARISON

| | Z-Score | Min-Max |
|---|---|---|
| JPM | 286.9 | 23.5 |
| FORD | 2437.1 | 43.9 |
| YHOO | 312.8 | 53.6 |
| Average | 1012.3 | 32.96 |

The min–max normalization procedure did not generalize as well as the z-score despite the good results in the other experiments based on the same stock data, as shown in Table V.

*B. Resultant Model*

After determining the best parameters, the most generalizable normalization method, and the training dataset, the resultant model was finally met. The model has five selected features. Its TWs are presented in Table VI.

TABLE VI. SELECTED TIME WINDOW

| Feature | TW |
|---|---|
| Stochastic %K | 11 |
| ADX | 6 |
| PSY | 14 |
| CCI | 7 |
| DI- | 9 |

*C. Stress Testing and Up-trend Rally Simulation*

The first test used to evaluate a model is a stress test, which is performed by financial analysts to examine a company's financial condition in a market stress situation. In the current study, the model was tested on data related to the market's financial crisis of 2008 and its partial recovery. The period begins on February 1, 2008 and ends on May 31, 2009. In Figure 4 and Table VII, the results of the stress test using GE, Pfizer, and Google stocks with US$100,000.00 of starting equity are shown and compared with the market default outcome called buy and hold (B&H). This outcome is the same as buying on the first day and selling on the last day of the period. As the model has the capability to avoid great market falls, recovery is achieved and good profits are obtained despite the crisis.

TABLE VII. RR AND MAX DRAWDOWN FOR THE STRESS TEST

| | RR [%] | MDD [%] | RR B&H [%] | MDD B&H [%] |
|---|---|---|---|---|
| GE | 118.9 | -25.68 | -58.65 | -81.37 |
| Pfizer | 104.1 | -14.49 | -29.22 | -46.28 |
| Google | 106 | -18.57 | -15.78 | -56.73 |

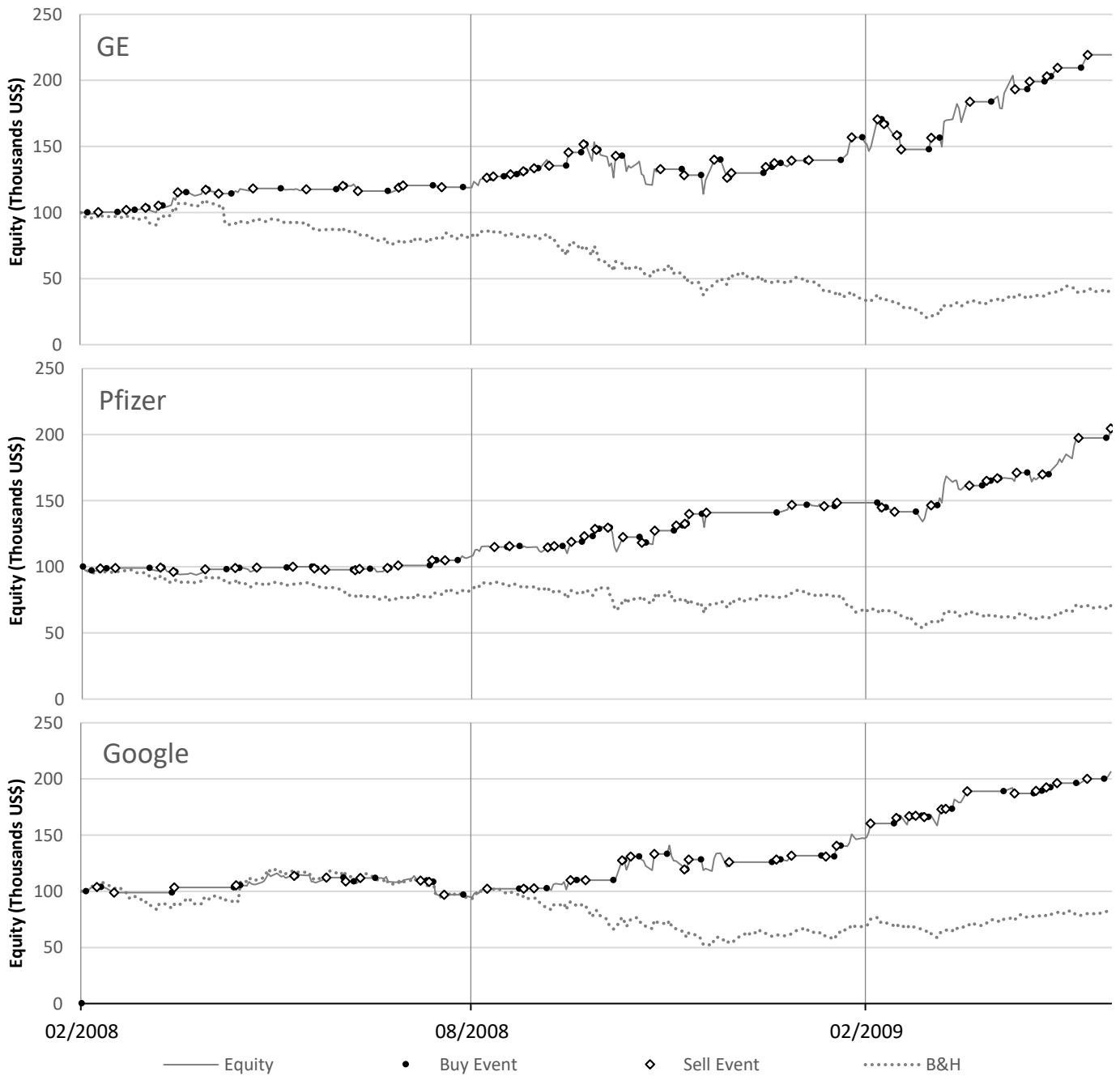

Fig. 4. Stress test.

To compare the efficacy of optimized TW SVM model, in Table VIII is presented the results obtained using default TW values, presented in Table III, without the proposed optimization of the TW.

TABLE VIII. DEFAULT TW STRESS TEST TRADING RESULTS

| Stock  | RR [%] | MDD [%] | RR B&H [%] | MDD B&H [%] |
|--------|--------|---------|------------|-------------|
| GE     | 0.3    | -46.7   | -58.65     | -81.37      |
| Pfizer | 26.2   | -22.2   | -29.22     | -46.28      |
| Google | 49.8   | -25.4   | -15.78     | -56.73      |

The results using the default TW are better than those suing the B&H. However, the results of GATWO are even better, as presented in Table VII, than those without the TW optimization. The *RR* of the non-optimized model is 25.43% on average, which is better than that of B&H, but the model using the TW optimized by GATWO achieves 109.66% on average.

After the stress test, the model was applied to stocks that showed a considerable up-trend rally. The stock market in this period was characterized by a subsequent appreciation of stock values. For this test, Apple (AAPL) stock from January 1, 2014 to January 31, 2015 and Microsoft stock from July 1, 2013 to

August 31, 2014 were used. As presented in Figure 5 and Table IX, the results of this test show a small margin of profit compared with the B&H strategy. Nevertheless, the test serves as a means to analyze whether the model overlooks any trading opportunities. As shown in Figure 5, the model sometimes does overlook some opportunities, but it also avoids some market falls. In both cases, the model outperforms the B&H strategy.

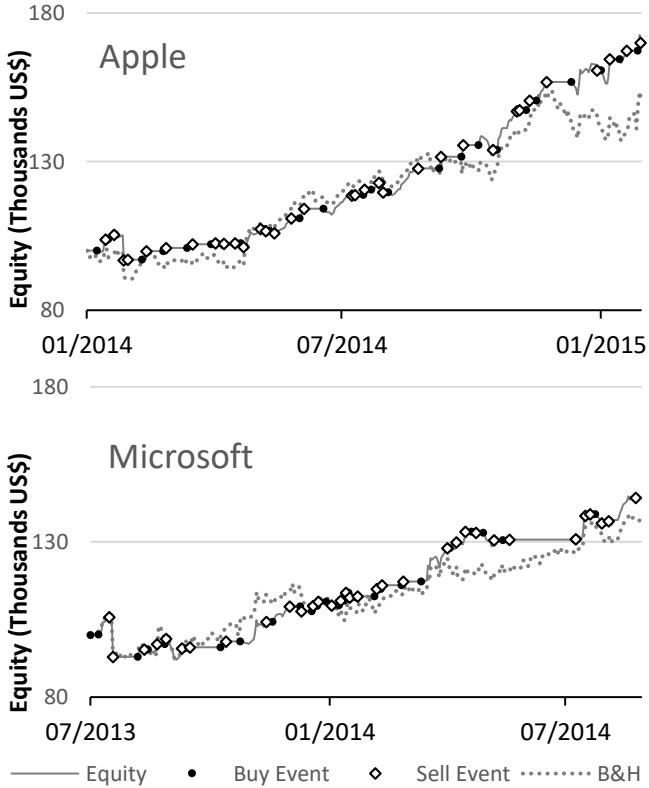

Fig. 5. Stress test.

TABLE IX. TRADING RESULTS BETWEEN TWO UP-TREND STOCK DATA

| Stock | RR[%] | MDD[%] | RR B&H [%] | MDD B&H [%] |
|---|---|---|---|---|
| AAPL | 69.8 | -8.93 | 51.4 | -10.93 |
| MSFT | 43.9 | -13 | 38.7 | -13.51 |

This trained model buys or sells stocks in 27.5% of all instances. In the remaining 72.5% of all instances, the trading model holds, waiting for a change in the classifier.

*D. State of the Art Comparison*

Other classifiers trained to determine the turning points are used to classify only a few trades at less than 10%, such as Chang's [12] intelligent piecewise linear representation (IPLR) and Li's [11] turning point prediction framework (TPP). This type of model overlooks some short-term trading opportunities. To overcome this limitation, the LG-Trader of Ng [3] was developed with three classes, namely, buy, sell, and hold, and the class distributions were balanced among the classes, with a penalty for the hold class. However, as presented in Figures 4 and 5, holding in a trend results in a better outcome. The moment when it is better to wait and hold for some time is extracted by the model from some features that were built on the basis of a trending detection method.

The proposed model was run with the same stock datasets, namely, IPLR, TPP, and LG-Trader, to compare the models. The results for these stocks are presented in Tables X and XI.

TABLE X. IPLR AND LG-TRADER RESULTS AND COMPARISON

| Stock | B&H | IPLR | LG-Trader | GATWO |
|---|---|---|---|---|
| UMC | -4.71 | 95 | 367.18 | 66.2 |
| AUO | 16.49 | 146 | 188.33 | 41.77 |
| COMPAL | 2.4 | 67 | 89.98 | 39.3 |

TABLE XI. TPP-FRAMEWORK AND LG-TRADER COMPARISON

| Stock | B&H | TPP | LG-Trader | GATWO |
|---|---|---|---|---|
| TESCO | 6.37 | 11.63 | 13.09 | 17.9 |

In addition, this model was compared with the VW-SVM trading engine of Żbikowski [13]. As in our model, the VW-SVM model was trained with a binary class, which classifies the trend situation, using the classification output to buy, hold, or sell. The VW-SVM model achieved good results when the model was trained with a delay classification. Żbikowski [13] used five days instead of one day ahead for classification in his best resultant model, and his experiments were conducted using 20 stocks and a wide time period of January 1, 2003 to October 21, 2013. The study used QCOM, GE, and MCD stocks. The comparison of these stocks is shown in Table VII).

TABLE XII. COMPARISON OF ŻBIKOWSKI'S VW-SVM TRADING RESULTS

| Stock | VW-SVM | | GATWO | |
|---|---|---|---|---|
| | RR [%] | MDD [%] | RR [%] | MDD [%] |
| QCOM | 68.94 | -49.14 | 26523.3 | -16.88 |
| GE | 1.01 | -31.64 | 11121.3 | -25.68 |
| MCD | 140.2 | -22.97 | 3561.2 | -18.25 |

V. CONCLUSION

The experimental results showed that the novel optimization of TW and the embedded FS in GATWO for predicting short-term trends in the stock market could lead to the creation of a profitable trading strategy. The resultant model was strong for down-trends, resisted market falls, and did not overlook good trading opportunities. Moreover, the several techniques applied for model generalization exhibited a strong generalization capability. The presented results indicated that the final model could be run with any other stocks without training the model.

# IEEE COPYRIGHT AND CONSENT FORM

To ensure uniformity of treatment among all contributors, other forms may not be substituted for this form, nor may any wording of the form be changed. This form is intended for original material submitted to the IEEE and must accompany any such material in order to be published by the IEEE. Please read the form carefully and keep a copy for your files.

A generalized financial time series forecasting model based on automatic feature engineering using genetic algorithms and support vector machine
Norberto Ritzmann Junior and Julio Cesar Nievola
2018 International Joint Conference on Neural Networks (IJCNN)

## COPYRIGHT TRANSFER

The undersigned hereby assigns to The Institute of Electrical and Electronics Engineers, Incorporated (the "IEEE") all rights under copyright that may exist in and to: (a) the Work, including any revised or expanded derivative works submitted to the IEEE by the undersigned based on the Work; and (b) any associated written or multimedia components or other enhancements accompanying the Work.

## GENERAL TERMS

1. The undersigned represents that he/she has the power and authority to make and execute this form.
2. The undersigned agrees to indemnify and hold harmless the IEEE from any damage or expense that may arise in the event of a breach of any of the warranties set forth above.
3. The undersigned agrees that publication with IEEE is subject to the policies and procedures of the IEEE PSPB Operations Manual.
4. In the event the above work is not accepted and published by the IEEE or is withdrawn by the author(s) before acceptance by the IEEE, the foregoing copyright transfer shall be null and void. In this case, IEEE will retain a copy of the manuscript for internal administrative/record-keeping purposes.
5. For jointly authored Works, all joint authors should sign, or one of the authors should sign as authorized agent for the others.
6. The author hereby warrants that the Work and Presentation (collectively, the "Materials") are original and that he/she is the author of the Materials. To the extent the Materials incorporate text passages, figures, data or other material from the works of others, the author has obtained any necessary permissions. Where necessary, the author has obtained all third party permissions and consents to grant the license above and has provided copies of such permissions and consents to IEEE

**You have indicated that you DO wish to have video/audio recordings made of your conference presentation under terms and conditions set forth in "Consent and Release."**

## CONSENT AND RELEASE

1. In the event the author makes a presentation based upon the Work at a conference hosted or sponsored in whole or in part by the IEEE, the author, in consideration for his/her participation in the conference, hereby grants the IEEE the unlimited, worldwide, irrevocable permission to use, distribute, publish, license, exhibit, record, digitize, broadcast, reproduce and archive, in any format or medium, whether now known or hereafter developed: (a) his/her presentation and comments at the conference; (b) any written materials or multimedia files used in connection with his/her presentation; and (c) any recorded interviews of him/her (collectively, the "Presentation"). The permission granted includes the transcription and reproduction of the Presentation for inclusion in products sold or distributed by IEEE and live or recorded broadcast of the Presentation during or after the conference.
2. In connection with the permission granted in Section 1, the author hereby grants IEEE the unlimited, worldwide, irrevocable right to use his/her name, picture, likeness, voice and biographical information as part of the advertisement, distribution and sale of products incorporating the Work or Presentation, and releases IEEE from any claim based on right of privacy or publicity.

BY TYPING IN YOUR FULL NAME BELOW AND CLICKING THE SUBMIT BUTTON, YOU CERTIFY THAT SUCH ACTION CONSTITUTES YOUR ELECTRONIC SIGNATURE TO THIS FORM IN ACCORDANCE WITH UNITED STATES LAW, WHICH AUTHORIZES ELECTRONIC SIGNATURE BY AUTHENTICATED REQUEST FROM A USER OVER THE INTERNET AS A VALID SUBSTITUTE FOR A WRITTEN SIGNATURE.

| Norberto Ritzmann Junior | 25-04-2018 |
|---|---|
| **Signature** | **Date (dd-mm-yyyy)** |

## Information for Authors

**AUTHOR RESPONSIBILITIES**

The IEEE distributes its technical publications throughout the world and wants to ensure that the material submitted to its publications is properly available to the readership of those publications. Authors must ensure that their Work meets the requirements as stated in section 8.2.1 of the IEEE PSPB Operations Manual, including provisions covering originality, authorship, author responsibilities and author misconduct. More information on IEEE's publishing policies may be found at http://www.ieee.org/publications_standards/publications/rights/authorrightsresponsibilities.html Authors are advised especially of IEEE PSPB Operations Manual section 8.2.1.B12: "It is the responsibility of the authors, not the IEEE, to determine whether disclosure of their material requires the prior consent of other parties and, if so, to obtain it." Authors are also advised of IEEE PSPB Operations Manual section 8.1.1B: "Statements and opinions given in work published by the IEEE are the expression of the authors."

**RETAINED RIGHTS/TERMS AND CONDITIONS**
- Authors/employers retain all proprietary rights in any process, procedure, or article of manufacture described in the Work.
- Authors/employers may reproduce or authorize others to reproduce the Work, material extracted verbatim from the Work, or derivative works for the author's personal use or for company use, provided that the source and the IEEE copyright notice are indicated, the copies are not used in any way that implies IEEE endorsement of a product or service of any employer, and the copies themselves are not offered for sale.
- Although authors are permitted to re-use all or portions of the Work in other works, this does not include granting third-party requests for reprinting, republishing, or other types of re-use.The IEEE Intellectual Property Rights office must handle all such third-party requests.
- Authors whose work was performed under a grant from a government funding agency are free to fulfill any deposit mandates from that funding agency.

**AUTHOR ONLINE USE**
- **Personal Servers**. Authors and/or their employers shall have the right to post the accepted version of IEEE-copyrighted articles on their own personal servers or the servers of their institutions or employers without permission from IEEE, provided that the posted version includes a prominently displayed IEEE copyright notice and, when published, a full citation to the original IEEE publication, including a link to the article abstract in IEEE Xplore. Authors shall not post the final, published versions of their papers.
- **Classroom or Internal Training Use.** An author is expressly permitted to post any portion of the accepted version of his/her own IEEE-copyrighted articles on the author's personal web site or the servers of the author's institution or company in connection with the author's teaching, training, or work responsibilities, provided that the appropriate copyright, credit, and reuse notices appear prominently with the posted material. Examples of permitted uses are lecture materials, course packs, e-reserves, conference presentations, or in-house training courses.
- **Electronic Preprints.** Before submitting an article to an IEEE publication, authors frequently post their manuscripts to their own web site, their employer's site, or to another server that invites constructive comment from colleagues. Upon submission of an article to IEEE, an author is required to transfer copyright in the article to IEEE, and the author must update any previously posted version of the article with a prominently displayed IEEE copyright notice. Upon publication of an article by the IEEE, the author must replace any previously posted electronic versions of the article with either (1) the full citation to the IEEE work with a Digital Object Identifier (DOI) or link to the article abstract in IEEE Xplore, or (2) the accepted version only (not the IEEE-published version), including the IEEE copyright notice and full citation, with a link to the final, published article in IEEE Xplore.

**Questions about the submission of the form or manuscript must be sent to the publication's editor.**
**Please direct all questions about IEEE copyright policy to:**
**IEEE Intellectual Property Rights Office, copyrights@ieee.org, +1-732-562-3966**

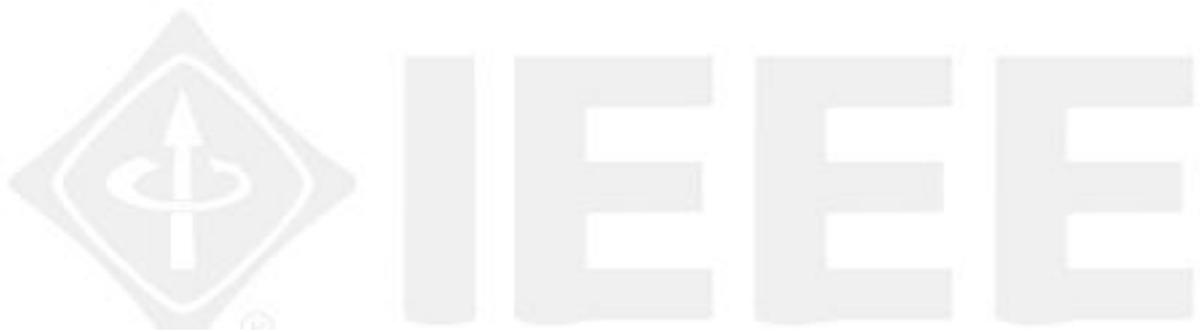